\pdfoutput=1
\documentclass[a4paper]{article}
\usepackage{INTERSPEECH2018}
\usepackage{threeparttable}

\title{Sampling strategies in Siamese Networks for unsupervised speech representation learning}
\name{Rachid Riad$^{1,2}$, Corentin Dancette$^1$, Julien Karadayi$^1$, Neil Zeghidour $^{1,3}$, Thomas Schatz$^{1,4,5}$, Emmanuel Dupoux$^{1,3}$}
\address{
  $^1$ CoML/ENS/CNRS/EHESS/INRIA/PSL Research University, Paris, France \\
  $^2$NPI/ENS/INSERM/UPEC, Cr\'eteil, France, 
  $^3$Facebook A.I. Research, Paris, France \\ 
  $^4$Department of Linguistics \& UMIACS, University of Maryland, College Park MD, USA \\
  $^5$Department of Linguistics, Massachusetts Institute of Technology, Cambridge MA, USA}
\email{\{emmanuel.dupoux,riadrachid3,julien.karadayi,corentin.dancette, neil.zeghidour, thomas.schatz.1986\}@gmail.com}

\begin{document}

\maketitle
%

% The maximum number of pages is 5. The 5\textsuperscript{th} page may be used exclusively for references. The references should begin on an earlier page immediately after the Acknowledgements section, and continue onto the 5\textsuperscript{th} page. If no space is available on an earlier page, then the references may begin on the 5\textsuperscript{th} page.

\begin{abstract}
	% This is the abstract
	%  How to build representation of speech sounds which is robust to within- and between-talker variation in a weakly-supervised or unsupervised setting? 
	Recent studies have investigated siamese network architectures for learning invariant speech representations using same-different side information at the word level. Here we investigate systematically an often ignored component of siamese networks: the sampling procedure (how pairs of same vs. different tokens are selected). We show that sampling strategies taking into account Zipf's Law, the distribution of speakers and the proportions of same and different pairs of words significantly impact the performance of the network. In particular, we show that word frequency compression improves learning across a large range of variations in number of training pairs. This effect does not apply to the same extent to the fully unsupervised setting, where the pairs of same-different words are obtained by spoken term discovery. We apply these results to pairs of words discovered using an unsupervised algorithm and show an improvement on state-of-the-art in unsupervised representation learning using siamese networks.
\end{abstract}
\section{Introduction}

Current speech and language technologies based on Deep Neural Networks (DNNs) \cite{hinton2012deep} require large quantities of transcribed data and additional linguistic resources (phonetic dictionary, transcribed data). 
Yet, for many languages in the world, such resources are not available and gathering them would be very difficult due to a lack of stable and widespread orthography \cite{adda2016breaking}.  

The goal of Zero-resource technologies is to build speech and language systems in an unknown language by using only raw speech data \cite{versteegh2015zero}. The Zero Resource challenges (2015 and 2017) focused on discovering invariant sub-word representations (Track 1) and audio terms (Track 2) in an unsupervised fashion. Several teams have proposed to use terms discovered in Track 2 to provide DNNs with pairs of same versus different words as a form of weak or self supervision for Track 1:  correspondence auto-encoders \cite{kamper2015unsupervised, renshaw2015comparison}, siamese networks \cite{thiolliere2015hybrid,zeghidour2016deep}.

This paper extends and complements the ABnet Siamese network architecture proposed by \cite{synnaeve2014phonetics, thiolliere2015hybrid} for the sub-word modelling task. DNN contributions typically focus on novel architectures or objective functions. Here, we study an often overlooked component of Siamese networks: the sampling procedure which chooses the set of pairs of same versus different tokens. 
To assess how each parameter contributes to the algorithm performance, we conduct a comprehensive set of experiments with a large range of variations in one parameter, holding constant the quantity of available data and the other parameters. 
We find that frequency compression of the word types has a particularly important effect. This is congruent with other frequency compression techniques used in NLP, for instance in the computation of word embeddings (\textit{word2vec} \cite{mikolov2013distributed}). Besides, Levy et al. \cite{levy2015improving} reveals that the performance differences between word-embedding algorithms are due more to the choice of the hyper-parameters, than to the embedding algorithms themselves. 

In this study, we first show that, using gold word-level annotations on the Buckeye corpus, a flattened frequency range gives the best results on phonetic learning in a Siamese network. Then, we show that the hyper-parameters that worked best with gold annotations yield improvements in the zero-resource scenario (unsupervised pairs) as well. Specifically, they improve on the state-of-the-art obtained with siamese and auto-encoder architectures. 

\section{Methods}

We developed a new package abnet3\footnote{\tiny \url{https://github.com/bootphon/abnet3}} using the pytorch framework  \cite{paszke2017automatic}. The code is open-sourced (BSD 3-clause) and available on github, as is the code for the experiments for this paper\footnote{\tiny \url{https://github.com/Rachine/sampling_siamese2018}}.

\subsection{Data preparation}

For the weakly-supervised study, we use 4 subsets of the Buckeye \cite{pitt2005buckeye} dataset from the ZeroSpeech 2015 challenge \cite{versteegh2015zero} with, respectively, 1\%, 10\%, 50\%, and 100\% of the original data (see Table \ref{tab:stats_buckeye}). The original dataset is composed of American English casual conversations recorded in the laboratory, with no overlap, no speech noises, separated in two splits: 12 speakers for training and 2 speakers for test. A Voice Activity Detection file indicates the onset and offset of each utterance and enables to discard silence portions of each file. We use the orthographic transcription from word-level annotations to determine same and different pairs to train the siamese networks. 

\begin{table}[!ht]
	\caption{Statistics for the 4 Buckeye splits used for the weakly supervised training, the duration in minutes expressed the total amount of speech for training}
	\label{tab:stats_buckeye}
	\centering
	\begin{tabular}{l|c|c|c|c}
		\toprule
		               & Duration    & \#tokens & \#words & \#possible pairs  \\
		\midrule
		\textbf{1\%}   & $3.0$ min   & $1006$   & $355$   & $\sim 5.10^{5}  $ \\
		\textbf{10\%}  & $29.9$ min  & $7189$   & $1297$  & $\sim 2.10^{7} $  \\
		
		\textbf{50\%}  & $149.5$ min & $34912$  & $3112$  & $ \sim 6.10^{8}$  \\
		\textbf{100\%} & $299.1$ min & $69543$  & $4538$  & $\geq 2.10^9 $    \\

		\bottomrule
	\end{tabular}
\end{table}
In the fully unsupervised setting, we obtain pairs of same and different words from the Track 2 baseline of the 2015 ZeroSpeech challenge \cite{versteegh2015zero}: the Spoken Term Discovery system from \cite{jansen2010towards}. We use both the original files from the baseline, and a rerun of the algorithm with systematic variations on its similarity threshold parameter.

For the speech signal pre-processing, frames are taken every 10ms and each one is encoded by a 40 log-energy Mel-scale filterbank representing 25ms of speech (Hamming windowed), without deltas or delta-delta coefficients. The input to the Siamese network is a stack of 7 successive filterbank frames. The features are mean-variance normalized per file, using the VAD information. 

\subsection{ABnet}
A \textit{Siamese network} is a type of neural network architecture that is used for representation learning, initially introduced for signature verification  \cite{bromley1994signature}. It contains 2 subnetworks sharing the same architecture and weights. In our case, to obtain the training information, we use the lexicon of words to learn an embedding of speech sounds which is more \textit{representative} of the linguistic properties of the  signal at the sub-word level (phoneme structure) and invariant to non-linguistic ones (speaker ID, channel, etc). 
A \textit{token} $t$ is from a specific word type $w$ (ex: ``the'',``process'' etc.) pronounced by a specific speaker $s$. The input to the network during training is a pair of stacked frames of filterbank features $x_1$ and $x_2$ and we use as label $y = \mathds{1}(\{w_1 = w_2\})$. 
For pairs of identical words, we realign them at the frame level using the Dynamic Time Warping (DTW) algorithm \cite{sakoe1978dynamic}. Based on the alignment paths from the DTW algorithm, the sequences of the stacked frames are then presented as the entries of the siamese network. Dissimilar pairs are aligned along the shortest word, e.g. the longest word is trimmed. 
With these notions of similarity, we can learn a representation where the distance between the two outputs of the siamese network $e(x_1)$ and $e(x_2)$ try to respect as much as possible the local constraints between $x_1$ and $x_2$. To do so, ABnet is trained with the margin cosine loss function:

\begin{equation*}
	l_{\gamma}(x_1,x_2,y)=\left\{
	\begin{array}{@{}ll@{}}
		-\cos (e(x_1),e(x_2)),                & \text{if}\ y=1   \\
		\max (0,\cos (e(x_1),e(x_2))-\gamma), & \text{otherwise} 
	\end{array}\right.
\end{equation*} 

For a clear and fair comparison between the sampling procedures we fixed the network architecture and loss function as in \cite{thiolliere2015hybrid}. The subnetwork is composed of 2 hidden layers with 500 units, with the Sigmoid as non-linearity and a final embedding layer of 100 units. For regularization, we use the Batch Normalization technique \cite{ioffe2015batch}, with a loss margin $\gamma=0.5$.
All the experiments are carried using the Adam training procedure \cite{DBLP:journals/corr/KingmaB14} and early-stopping on a held-out validation set of $30\%$ of spoken words. We sample the validation set in the same way as the training set.

\subsection{Sampling}
The sampling strategy refers to the way pairs of tokens are fed to the Siamese network. Sampling every possible pairs of tokens becomes quickly intractable as the dataset grows (cf. Table \ref{tab:stats_buckeye}). 

There are four different possible configurations for a pair of word tokens $(t_1,t_2) $ : whether, or not, the tokens are from the same word type, $w_1 = w_2$. and whether, or not, the tokens are pronounced by the same speaker, $s_1 = s_2$. 

Each specific word type $w$ is characterized by the total number of  occurrences $n_w$ it has been spoken in the whole corpus.  Then, is deduced the frequency of appearances $f_w \propto n_w$, and  $r_w$ its frequency rank in the given corpus. We want to sample a pair of word tokens, in our framework we sample independently these 2 tokens. 
We define the probability to sample a specific token word type $w$ as a function of $n_w$. We introduce the function $\phi$ as the \textit{sampling compression function}:

\begin{equation}
	\mathbb{P}(w) = \frac{\phi(n_w)}{\sum\limits_{\forall w'}\phi(n_{w'})}
\end{equation} 

When a specific word type $w$ is selected according to these probabilities, a token $t$ is selected randomly from the specific word type $w$. The usual strategy to select pairs to train siamese networks is to randomly pick two tokens from the whole list of training tokens examples \cite{bromley1994signature, chopra2005learning,thiolliere2015hybrid}. In this framework, the sampling function corresponds $\phi: n \rightarrow n$. 
Yet, there is a puzzling phenomenon in human language, there exists an empirical law for the distribution of words, also known as the \textit{Zipf's law} \cite{zipf1935psycho}. Words types appear following a power law relationship between the frequency $f_w$ and the corresponding rank $r_w$: a few very high-frequency types account for almost all tokens in a natural corpus (most of them are function words such as ``the'',``a'',``it'', etc.) and there are many word types with a low frequency of appearances (``magret'',``duck'',``hectagon''). The frequency $f_t$  of type $t$ scales with its corresponding $r_t$ following a power law, with a parameter $\alpha$ depending on the language: 

\begin{equation*}
	f_w \propto \frac{1}{r_w^\alpha}, \alpha \approx 1
\end{equation*} 

One main effect on the training is the oversampling of word types with high frequency, and this is accentuated with the sampling of two tokens for the siamese. These frequent, usually monosyllabic, word types do not carry the necessary phonetic diversity to learn an embedding robust to rarer co-articulations, and rarer phones. 
To study and minimize this empirical linguistic trend, we will examine 4 other possibilities for the $\phi$ function that compress the word frequency type:
\begin{align*}
	  & \phi: n \rightarrow \sqrt[2]{n}, &\phi: n \rightarrow \sqrt[3]{n} \\
	  & \phi: n \rightarrow \log(1+n),   &\phi: n \rightarrow 1           
\end{align*}

The first two options minimize the effect of the Zipf's Law on the frequency, but the power law is kept.
The $\log$ option removes the power law distribution, yet it keeps a linear weighting as a function of the rank of the types. Finally with the last configuration, the word types are sampled uniformly.

Another important variation factor in speech realizations is the speaker identity. We expect that the learning of speech representations to take advantage of word pairs from different speakers, to generalize better to new ones, and improve the \textit{ABX performance}.

\begin{equation*}
	P^s_{-} = \frac{\# \text{Sampled pairs pronounced by different speakers}}{\# \text{Sampled pairs}}
\end{equation*}

Given the natural statistics of the dataset, the number of possible "different" pairs exceeds by a large margin the number of possible "same" pairs ($\sim 1\%$ of all token pairs for the Buckeye-100\%). The siamese loss is such that "Same" pairs are brought together in embedding space, and "Different" pairs are pulled apart. Should we reflect this statistic during the training, or eliminate it by presenting same and different pairs equally? We manipulate systematically the proportion of pairs from different word types fed to the network:

\begin{equation*}
	P^w_{-} = \frac{\# \text{Sampled pairs with non-matching word types}}{\# \text{Sampled pairs}}
\end{equation*}

\subsection{Evaluation with ABX tasks}
To test if the learned representations can separate phonetic categories, we use a minimal pair ABX discrimination task \cite{schatz2013evaluating, schatz2014evaluating}. It only requires to define a dissimilarity function $d$ between speech tokens, no external training algorithm is needed. We define the ABX-discriminability of category $x$ from category $y$ as the probability that $A$ and $X$ are further apart than $B$ and $X$ when $A$ and $X$ are from category $x$ and $B$ is from category $y$, according to a dissimilarity function $d$. Here, we focus on phone triplet minimal pairs: sequences of 3 phonemes that differ only in the central one (``beg''-``bag'', ``api''-``ati'', etc.). For the \textit{within-speaker task}, all the phones triplets belong to the same speaker (e.g. $A = beg_{T_1}, B = bag_{T_1}, X = bag'_{T_1}$)
Finally the scores for every pair of central phones are averaged and subtracted from 1 to yield the reported within-talker ABX error rate. For the \textit{across-speaker task}, $A$ and $B$ belong to the same speaker, and $X$ to a different one (e.g. $A = beg_{T_1}, B = bag_{T_1}, X = bag'_{T_2}$). The scores for a given minimal pair are first averaged across all of the pairs of speakers for which this contrast can be made. As above, the resulting scores are averaged over all contexts over all pairs of central phones and converted to an error rate.

\section{Results}

\subsection{Weakly supervised Learning}

\subsubsection{Sampling function $\phi$}

\begin{figure}
	\centering
	
	\includegraphics[width=\linewidth]{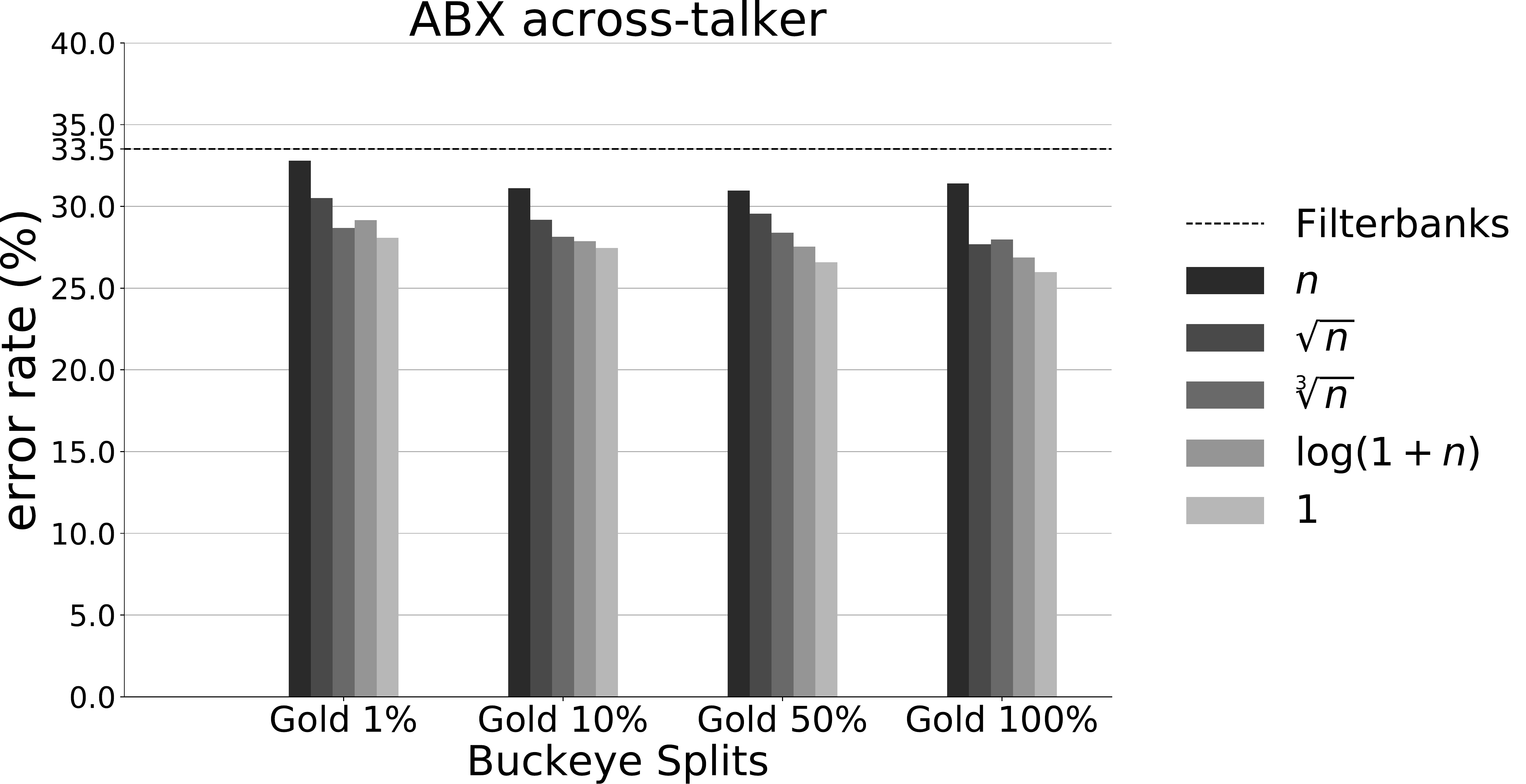}
	\caption{ABX across-speaker error rates  on test set with various sampling compression functions $\phi$ for the 4 different Buckeye splits used for weakly supervised training. Here, the proportions of pairs with different speakers $P^s_{-}$ and with different word types $P^w_{-}$  are kept fixed: $P^s_{-}=0.5$, $P^w_{-}=0.5$}
	\label{fig:sampling_functions}
	
\end{figure}

We first analyze the results for the \textit{sampling compression function} $\phi$ Figure \ref{fig:sampling_functions}. For all training datasets, we observe a similar  pattern for the performances on both tasks: the word frequency compression improves the learning and generalization. The result show that, compared to the raw filterbank features baseline, all the trained ABnet networks improve the scores on the phoneme discrimination tasks, even in the $1\%$ scenario. Yet, the improvement with the usual sampling scenario $\phi: n \rightarrow n$ is small in all 4 training datasets. The optimal function for the within and across speaker task on all training configuration is the uniform function $\phi: n \rightarrow 1$. It yields substantial improvements over the raw filterbanks for ABX task across-speaker ( $5.6 $ absolute points and $16.8 \%$ relative improvement for the $1\%$-Buckeye training). The addition of data for these experiments improves the performance of the network, but not in a substantial way: the improvements from $1\%$-Buckeye to $100\%$-Buckeye, for $\phi: n \rightarrow 1$, is  $1.9 $ absolute points and $ 7.9\%$ relative. 
These results show that using frequency compression is clearly beneficial, and surprisingly adding more data is still advantageous but not as much as the choice of $\phi$. Renshaw et al. \cite{renshaw2015comparison}, found similar results with a correspondence auto-encoder, training with more training data did not yield improvements for their system.

\subsubsection{Proportion of pairs from different speakers $P_{-}^s$}

\begin{figure}[h]
	\centering
	\includegraphics[width=\linewidth]{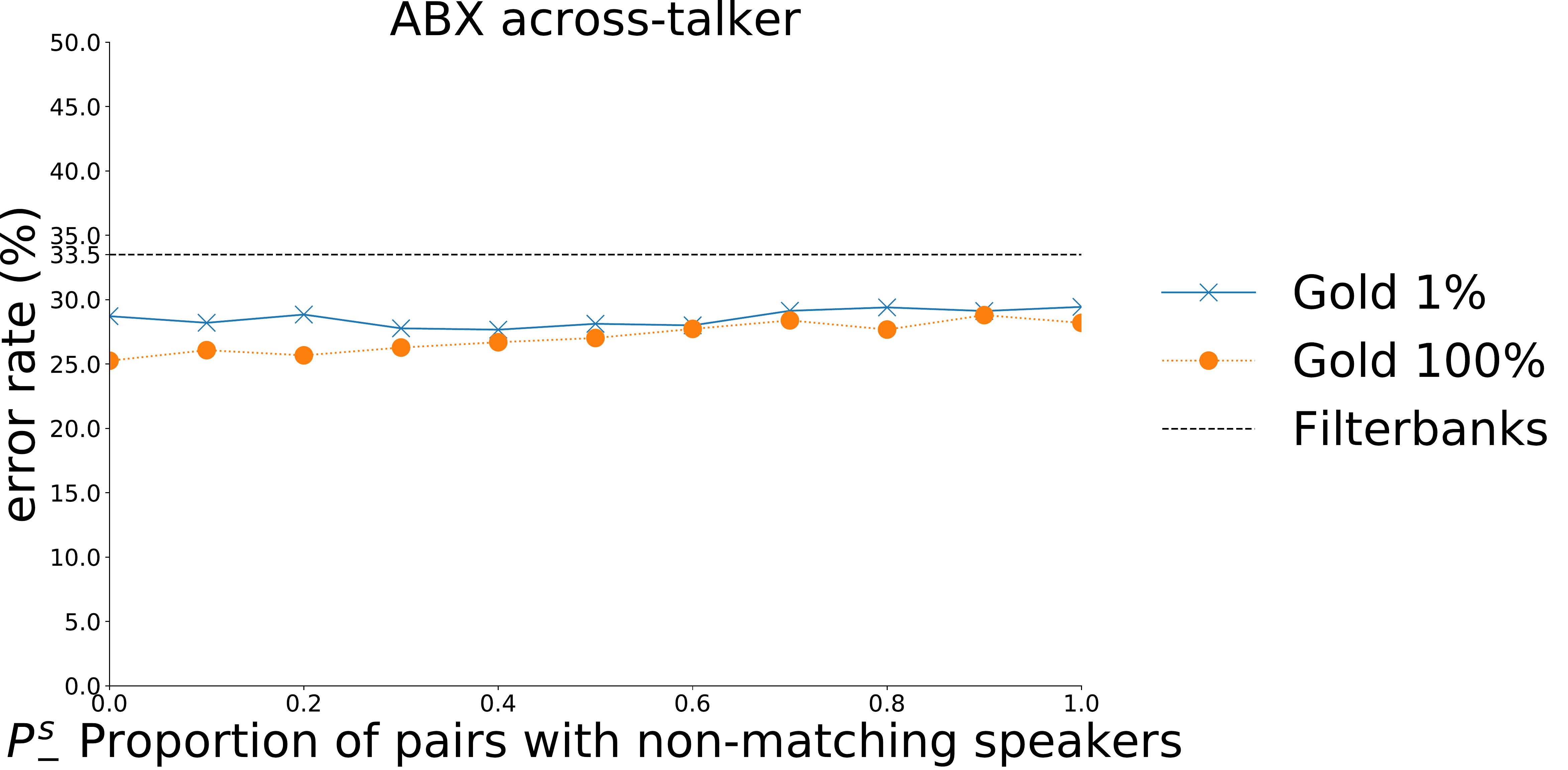}
	 
	\caption{Average ABX error rates across-speaker with various proportion pairs of different speakers $P^s_{-}$, with $\phi: n \rightarrow 1$ and $P^w_{-}=0.5$. }
	\label{fig:proportion_pairs_spk_across}
	
\end{figure}

We now look at the effect on the ABX performances of the proportion of pairs of words pronounced by two different speakers Figure \ref{fig:proportion_pairs_spk_across}.  We start from our best sampling function configuration so far $\phi: n \rightarrow 1$. We report on the graph only the two extreme training settings. The variations for the 4 different training splits are similar, and still witness a positive effect with additional data on the siamese network performances. Counter-intuitively, the performances on the ABX tasks does not take advantage of pairs from different speakers. It even shows a tendency to increase the ABX error rate: for the $100\%$-Buckeye we witness an augmentation of the ABX error-rate (2.9 points and  $11.6\%$ relative) between $P_{-}^s=0$ and $P_{-}^s=1$. One of our hypothesis on this surprising effect, might be the poor performance of the DTW alignment algorithm directly on raw filterbanks features of tokens from 2 different speakers.

\subsubsection{Proportion of pairs with different word types $P_{-}^w$}

\begin{figure}[h]
	\centering
	\includegraphics[width=\linewidth]{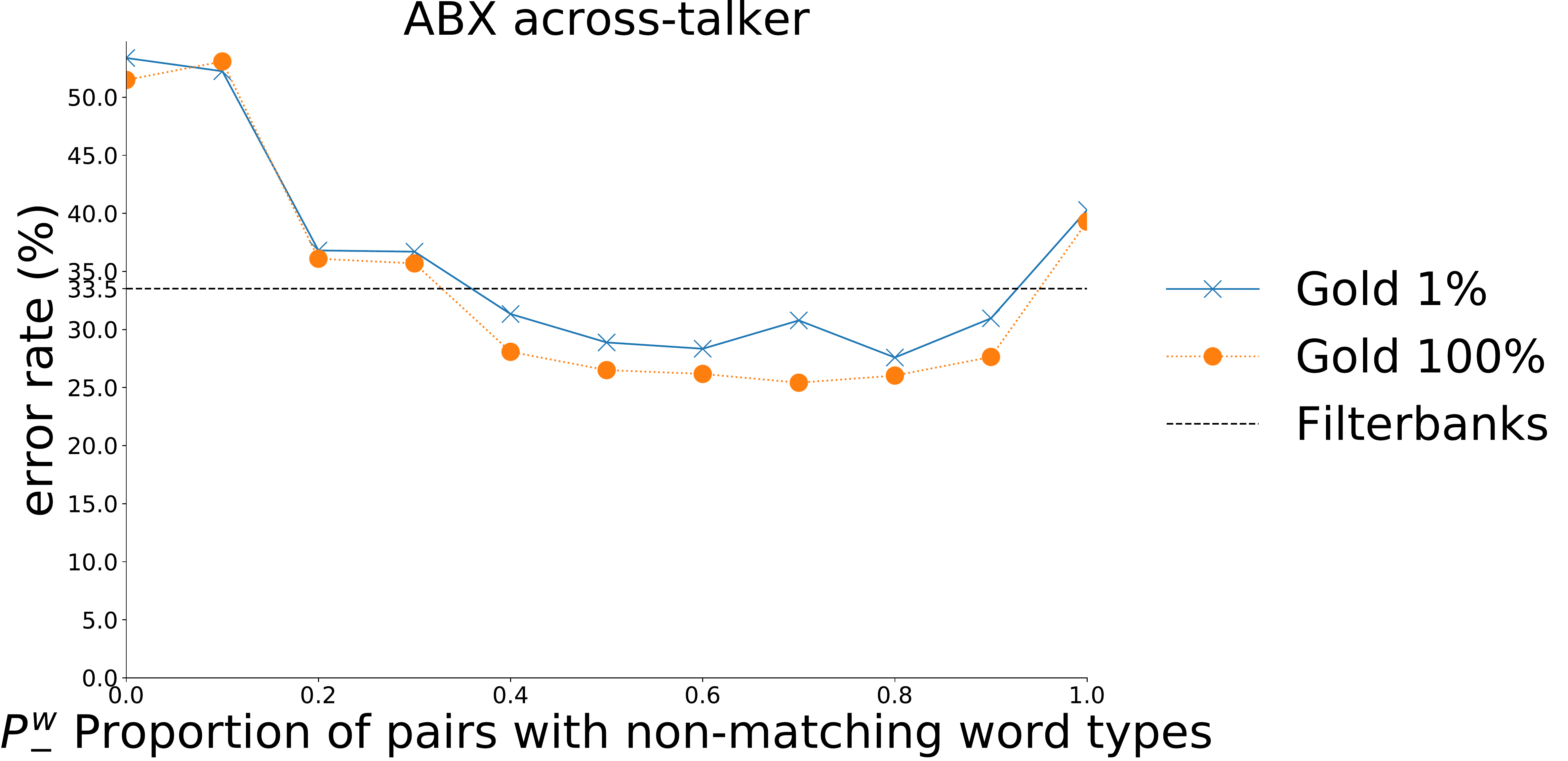}
	\caption{ Average ABX error rates across-speaker with various proportion pairs with different word types $P^w_{-}$ , where $\phi: n \rightarrow 1$ and $P^s_{-}=0.5$}
	\label{fig:proportion_pairs_type_across}
\end{figure}

We next study the influence of the proportion of pairs from different word types $P^w_{-}$ Figure \ref{fig:proportion_pairs_type_across}. In all training scenarios, to privilege either only the positive or the negative examples is not the solution. 
For the different training splits, the optimal number for $P_{-}^w$ is either $0.7$ or $0.8$ in the within and across speaker ABX task. We do not observe a symmetric influence of the positive and negative examples, but it is necessary to keep the same and different pairs. The results collapsed, if the siamese network is provided only with positive labels to match: the network will tend to map all speech tokens to the same vector point and the discriminability is at chance level. 

\subsection{Applications to fully unsupervised setting}
\subsubsection{ZeroSpeech 2015 challenge}

Now, we transfer the findings about sampling from the weakly supervised setting, to the fully unsupervised setting. We report  in Table \ref{tab:unsupervised-magic} our results for the two ZeroSpeech 2015\cite{versteegh2015zero} corpus: the same subset of the Buckeye Corpus as earlier and a subset of the NCHLT corpus of Xitsonga \cite{de2014smartphone}.  To train our siamese networks, we use as \cite{thiolliere2015hybrid}, the top-down information from the baseline for the Track 2 (Spoken Term Discovery) of the ZeroSpeech 2015 challenge from \cite{jansen2010towards}. The resulting clusters are not perfect, whereas we had perfect clusters in our previous analysis.

\begin{table}[!ht]
		\caption{ABX discriminability results for the ZeroSpeech2015 datasets. The best error rates for each conditions for siamese architectures are in \textbf{bold}. The best error rates for each conditions overall are \underline{underlined}.}
		\label{tab:unsupervised-magic}
		\centering
		\resizebox{\textwidth}{!}{
			\begin{tabular}{lccccc}
				\toprule
				Models & \multicolumn{2}{c}{English} &  \multicolumn{2}{c}{Xitsonga}    \\
				                                          & within                    & across           & within                   & across                    \\
				\midrule
				baseline (MFCC)                           & 15.6                      & 28.1             & 19.1                     & 33.8                      \\
				supervised topline (HMM-GMM)              & 12.1                      & 16.0             & 04.5                     & 03.5                      \\
				\midrule
				Our ABnet with $P^w_{-}=0.7, P^s_{-}=0, \phi : n \rightarrow 1 $ 
				                                          & \underline{\textbf{10.4}} & 17.2             & \textbf{9.4} & \textbf{15.2} \\
				
				\midrule  
				CAE, Renshaw et al. \cite{renshaw2015comparison} 
				                                          & 13.5                      & 21.1             & 11.9                     & 19.3                      \\
				ABnet, Thioli\`ere et al. \cite{thiolliere2015hybrid}  
				                                          & 12.0                      & 17.9             & 11.7                     & 16.6                      \\
				ScatABnet, Zeghidour et al. \cite{zeghidour2016deep} 
				                                          & 11.0                      & \textbf{17}      & 12.0                     & 15.8                      \\
				DPGMM Chen et al. \cite{chen2015parallel} & 10.8                      & 16.3 & 9.6                      & 17.2                      \\
				DPGMM+PLP+bestLDA+DPGMM Heck et al. \cite{heck2016unsupervised} & 10.6                      & \underline{16.0} & \underline{8.0}                      & \underline{12.6}                      \\
				\bottomrule
			\end{tabular}}
\end{table}

In Thioli\`ere et al. \cite{thiolliere2015hybrid} the sampling is done with : $P^w_{-} = P^s_{-} = 0.5$, and $\phi = n \rightarrow n$. This gives us a baseline to compare our sampling method improvements with our own implementation of siamese networks. 
% We also reported Thioliere et al.\cite{thiolliere2015hybrid} and  Zeghidour et al.\cite{zeghidour2016deep} results for comparison with previous siamese networks architectures. In bold are our best results.

First, the ``discovered'' clusters -- obtained from spoken term discovery system -- don't follow the Zipf's law like the gold clusters. 
This difference of distributions diminishes the impact of the sampling compression function $\phi$.

We matched state-of-the-art for this challenge only on the ABX task  within-speaker for the Buckeye, otherwise the modified DPGMM algorithm proposed by Heck et al. stays the best submissions for the 2015 ZeroSpeech challenge. 
\subsubsection{Spoken Term discovery - DTW-threshold $\delta$}

Finally, we study the influence of the DTW-threshold $\delta$ used in the spoken discovery system on the phonetic discriminability of siamese networks. We start again from our best finding from weakly supervised learning. The clusters found by the Jansen et al. \cite{jansen2010towards} system are very sensitive to this parameter with a trade-off between the \textit{Coverage} and the \textit{Normalized Edit Distance} (NED) introduced by \cite{ludusan2014bridging}. 

\begin{table}[!ht]
	\caption{Number of found clusters, NED, Coverage, ABX discriminability results with our ABnet with $P^w_{-}=0.7, P^s_{-}=0, \phi : n \rightarrow 1 $, for the ZeroSpeech2015 Buckeye for various DTW-thresholds $\delta$ in the Jansen et al. \cite{jansen2010towards} STD system. 
	The best results for each metric are in \textbf{bold}.}
	\label{tab:dtw_thresh}
	\centering
	\begin{tabular}{l|c|c|c|c}
		\toprule
		 $\delta$ & \#clusters & NED & Coverage & ABX across \\
		\midrule
        0.82 &  27,770 & 0.792 & \textbf{0.541} &  18.2    \\
        0.83 &  27,758 & 0.792 & 0.541          &  18.1   \\
        0.84 &  27,600 & 0.789 & 0.541          &   18.4   \\
        0.85 &  26,466 & 0.76  & 0.54           &   18.4      \\
        0.86 &  22,627 & 0.711 & 0.527          &  18.2  \\
        0.87 &  16,108 & 0.569 & 0.485          &  18.2    \\
        0.88 &  9,853  & 0.442 & 0.394          &   17.7   \\
        0.89 &  5,481  & 0.309 & 0.282          &  \textbf{17.6}     \\
        0.90 &  2,846  & 0.228 & 0.182          &   17.9  \\
        0.91 &  1,286  & 0.179 & 0.109          &  18.6      \\
        0.92 &  468    & \textbf{0.179} & 0.058  &  19.2      \\
		\bottomrule
	\end{tabular}
\end{table}

We find that ABnet is getting good results across the various outputs of the STD system shown in Table \ref{tab:dtw_thresh} and improves over the filterbanks results in all cases. Obtaining more data with the STD system involves a loss in words quality. In contrast with the weakly supervised setting, there is an \textit{optimal trade-off} between the amount and quality of discovered words for the sub-word modelling task with siamese networks.

\section{Conclusions and Future work}
 We presented a systematic study of the sampling component in siamese networks. In the weakly-supervised setting, we established that the word frequency compression had an important impact on the discriminability performances. We also found that optimal proportions of pairs with different types and speakers are not the ones usually used in siamese networks. 
We transferred the best parameters to the unsupervised setting to compare our results to the 2015 Zero Resource challenge submissions. It lead to improvements over the previous neural networks architectures, yet the Gaussian mixture methods (DPGMM) remain the state-of-the-art in the phonetic discriminability task.
In the future, we will study in the same systematic way the influence of sampling in the fully unsupervised setting. We will then try to leverage the better discriminability of our representations obtained with ABnet to improve the spoken term discovery, which relies on frame-level discrimination to find pairs of similar words.
Besides, power law distributions  are endemic in natural language tasks. It would be interesting to extend this principle to other tasks (for instance, language modeling).

\section{Acknowledgements}
The team's project is funded by the European Research Council (ERC-2011-AdG-295810 BOOTPHON), the Agence Nationale pour la Recherche (ANR-10-LABX-0087 IEC, ANR-10-IDEX-0001-02 PSL* ), Almerys (industrial chair Data Science and Security), Facebook AI Research (Doctoral research contract), Microsoft Research (joint MSR-INRIA center) and a Google Award Grant.

\newpage

\bibliographystyle{IEEEtran}

\bibliography{paper.bib}

\end{document}